\documentclass[letterpaper,twocolumn,10pt]{article}
\usepackage{usenix2019_v3}
\usepackage{float}

\usepackage{tikz}
\usepackage{amsmath}
\usepackage{enumitem}

\usepackage{filecontents}

\begin{document}

\date{}

\title{\Large \bf CS294-196 Decentralized Systems Project Report:\\
  Federated Learning, but Fully Decentralized}

\author{
{\rm Ram Kripa}\\
ram.m.kripa@berkeley.edu
\and
{\rm Andy Zou}\\
andyzou\_jiaming@berkeley.edu
\and
{\rm Ryan Jia}\\
ryanjia@berkeley.edu
\and
{\rm Kenny Huang}\\
kennyhuang9@berkeley.edu
}

\maketitle


\begin{abstract}
Federated Learning systems use a centralized server to aggregate model updates. This is a bandwidth and resource-heavy constraint and exposes the system to privacy concerns. We instead implement a peer to peer learning system in which nodes train on their own data and periodically perform a weighted average of their parameters with that of their peers according to a learned trust matrix. So far, we have created a model client framework and have been using this to run experiments on the proposed system using multiple virtual nodes which in reality exist on the same computer. We used this strategy as stated in Iteration 1 of our proposal to prove the concept of peer to peer learning with shared parameters. We now hope to run more experiments and build a more deployable real world system for the same.
\end{abstract}

\section{Introduction}
Conventional distributed machine learning systems often lack privacy because of the communication between individual nodes and the centralized server during model updates. The practice of aggregating parameter updates on the central server at once compromises security and privacy. Peer to peer federated learning systems improves upon it by updating nodes within some random cohort following some simple update rules, therefore completely taking away the need for the central aggregator. However, previously proposed systems were mostly theoretical and in simplified settings. They did not fully exploit the structures of local updates with a learned peer-to-peer update rule. We wish to fulfill this need for a peer-to-peer federated learning framework that is robust and provides good training times and accuracy.

\section{Background}
\subsection{Federated Learning}
Federated Learning differs from conventional distributed machine learning due to the very large number of clients, highly unbalanced and non-i.i.d. data available on each client, and relatively poor network connections. Since the unreliable and asymmetric connections pose a particular challenge to practical Federated Learning, another group of researchers focus on the relatively poor network connections. As one of the first papers that introduce Federated Learning, \cite{D} explores deeper into the field by looking into two ways to reduce the uplink communication costs: structured updates and sketched updates. 

In summary, structured updates directly learn an update from a restricted space that can be parameterized using a smaller number of variables. Sketched updates learn a full model update, then compress it before sending to the server. In order to understand these two approaches, let us first define the problem. Let $W \in \mathbb{R}^{d1 \times d2}$ be the learning model's parameters. In time  $t \geq 0$, the model $W_t$ is distributed to $n_t$ clients. These clients independently update the model based on their local data. Let the updated local models be $W^1_t, W^2_t, ... W^{n_t}_t$ and we define $H^i_t = W^i_t - W_t$ be the update client i needs to send back to the server. As a result, after collecting all the updates from client, the server performs the global update:
$$W_{t+1} = W_t + \alpha_t H_t$$   $$ H_t := \frac{1}{n_t}\sum_i H^i_t$$

\subsection{Structured Updates}
In structured update approach, the update matrix $H_t$ is restricted to have a pre-specified structure. There are 2 types of structured updates: low rank and random mask.

In low rank structured update, We enforce every update to local model to be a low rank matrix of rank at most k, where k is a given number. In particular, we express $H^i_t$ as the product of the 2 matrices. $H^i_t = A^i_t B^i_t$, where $A^i_t \in \mathbb{R}^{d_1 \times k}$ and $B^i_t \in \mathbb{R}^{k \times d_2}$. For each client i,  we generate 
$A^i_t$ randomly and consider a constant during a local training procedure, and we optimize only $B^i_t$. By doing so, we only send back a update matrix of at most rank k back to the server. Note that the construction of $A^i_t$ can be compressed as a random seed.

In random mask structured update, We restrict $H^i_t$ to be a sparse matrix. In order to do so, we apply a sparsity mask on $H^i_t$. The mask is random and pre-determined by a random seed. By doing so, each client only needs to send back non-zero entries back to the server.

\subsection{Sketched Updates}
Slightly different from the structured update, sketched updates compute the full $H^i_t$, then encodes $H^i_t$ in a compressed form back to the client. The authors introduce three approaches: subsampling, probabilistic quantization, and structured random rotations.

In subsampling, the idea is rather straightforward, instead of sending back the full  $H^i_t$, each client only is required to send back a random subset of scaled values of  $H^i_t$.  The server then averages the subsampled updates,producing the global. This can be done so that the average of the sampled updates is an unbiased estimator of the true average. Similar to the random mask structured update, the mask is randomized independently for each client in each round, and the mask itself can be stored as a synchronized seed.

In probabilistic quantization, we can compress or quantize each scalar to only 1 bit. Now consider a scalar $h_i$ where $i \in [1, d1 \times d2]$. let $h_{max} = max_i (h_i)$ and $h_{min} = min_i (h_i)$. The quantized scalar can then be expressed as:
$$
\tilde{h_i}= \begin{cases} 
    h_{max}  & \text{with probability} \frac{h_j - h_{min}}{h_{max} - h_{min}} \\
    h_{min}  & \text{with probability} \frac{h_{max} - h_{j}}{h_{max} - h_{min}}
    \end{cases}
$$

It is easy to see that this approach compresses the original update matrix 32 fold and is called 1-bit quantization. Moreover, $b$-bit quantizations for different $b$ values can also be applied.

Finally, structured random rotations should be applied along with probabilistic quantization. In particular, $b$-bit quantization work best when the scales are approximately equal across different dimensions. For example, when $max = 1$ and $min = -1$ and most of values are 0, the 1-bit quantization will lead to a large error. The paper proposes that applying a random rotation on $h$ before the quantization (multiplying $h$ by a random orthogonal matrix) solves this issue.

\subsection{GBoard}

Such techniques introduced above and variants were deployed in real life with stunning successes. Instead of storing the training data and performing machine learning on a centralized server, Google built highly secure and robust cloud infrastructures used for Federated Learning \cite{B}. This achieves superior privacy and decentralization where the data is only stored on individual devices in addition to the model which is also being trained on the same devices. 

Notably, the Federated Learning system works well when \quad 1. the task labels are collected during user interaction rather than manual labeling, \quad 2. the training data contains user sensitive information, \quad 3. the training data is too large to be collected with a central server \cite{A}.

An especially neat application currently is the Google Keyboard. After downloading the model from a central server, the local models are updated and used in real time for better experience. The only communication back to the server is a focused update which is being averaged with the feedback updates from other users. Then these updates are being decrypted and merged with the central model only when a big batch users have participated. This is called Federated Averaging and further enhances privacy by transmitting as minimal amount of information as possible in a short amount of time.

Specifically, a baseline model for query suggestion is trained offline on the server with traditional machine learning techniques by matching the user's input to a subset of the Google Knowledge Graph. Then a Long Short-Term Memory network is used to score potential query candidates. In order to augment the server-side model with personalized data, Google employs Federated Learning in its triggering model. This models serves as a feed filter and decides whether or not the query suggestion generated from the baseline model should be shown to the end user based on the training cache stored on the device. The results show that the incorporation of the logistic triggering model leads to a successful improvement in the click-through-rate.

\subsection{Blockchain-based Federated Learning}

By some examples shown above, federated learning (FL) is essential for the expansion of large-scale machine learning (ML) models, while at the same time strives to preserve the privacy of client’s local data. However, same as any algorithms, recent experiments illustrate that FL is not without its own caveats.

According to \cite{F}, FL could suffer from:
\begin{itemize}[leftmargin=20pt]
    \item Single point of failure. The central server (aggregator) is the most critical component of traditional FL algorithms. The entire training process would be sabotaged if the aggregator decides to intentionally perform dishonestly or is exposed to compromises. In other words, we want to have even more decentralized algorithms.
    \item Malicious clients and invalid parameter uploadings. Adversarial data attacks are one of the most common attacks towards ML models. FL does not have an intrinsic way to validate the data or client identities.
    \item The lack of incentives. Why would a client want to contribute the computation resources for… nothing? We don’t have an incentive framework in traditional FL.
\end{itemize}

It’s natural to see that blockchain infrastructure is capable of solving all the listed problems above. In \cite{F}, 3 different categories are proposed in terms of how the FL models combine with the blockchain backbone. All of the structures are called blockchain-based FL (BCFL). We restate them as below.

This way is the most entangled way between BC and FL:
\begin{center}
    \includegraphics[width=0.4\textwidth]{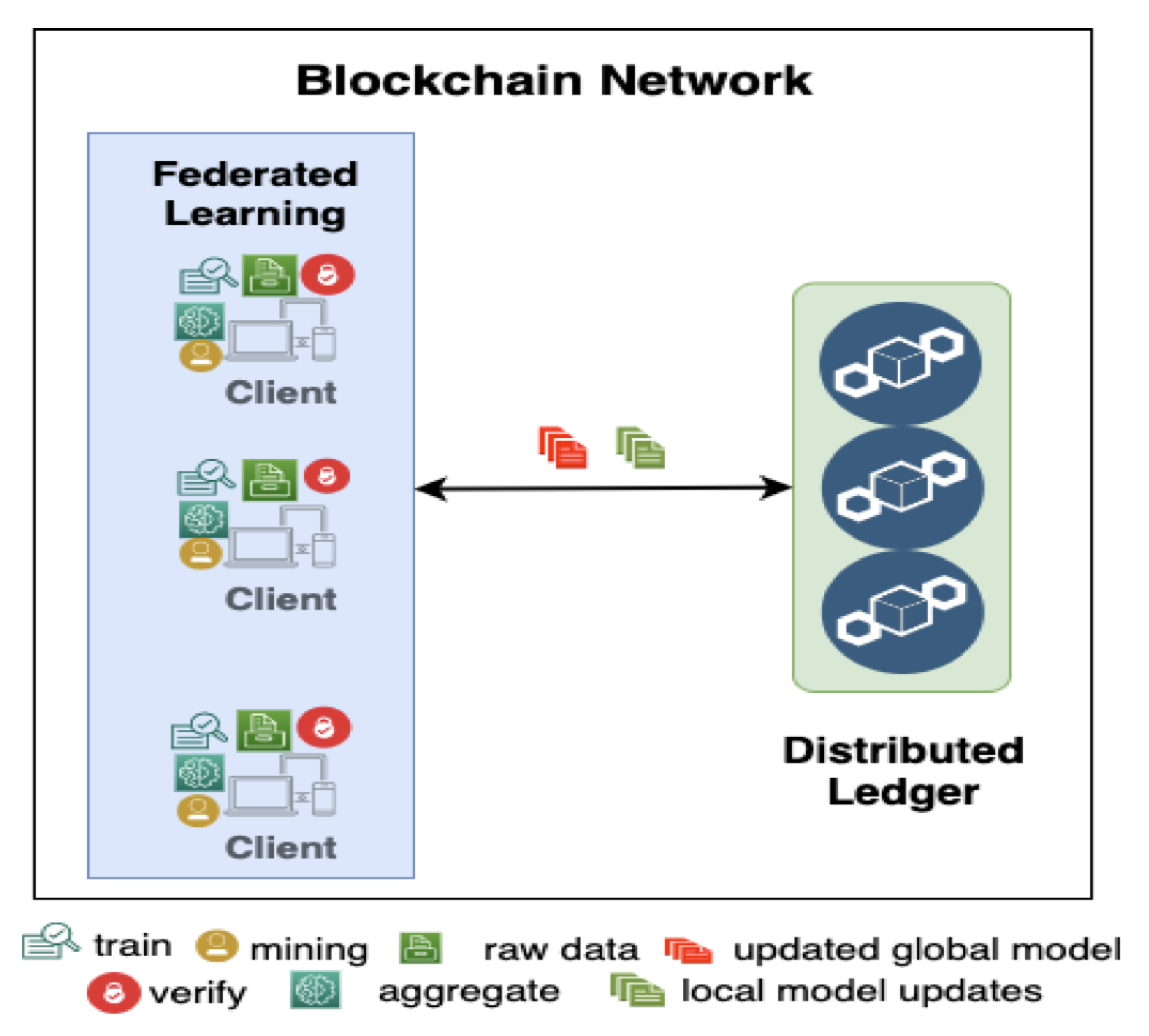}
\end{center}

This is referred to as the fully coupled BCFL (FuC-BCFL) in the paper. As illustrated, all the nodes in the network take on both the roles of model training and ledger block computation. The model parameters are saved on the ledger overall, and are verified by all competing nodes by construction, so it's resistant to malicious updating attack. All the nodes can participate in the global model aggregation, so that we could completely remove the need for a central aggregator, unlike the traditional FL. One concrete example protocol would be that the local model update could only be uploaded by the first node who compute a block \cite{G}. The downside to this approach, however, is also very obvious: since the communication bandwidth of BC is usually limited, training latency is expected, and more computational resources are required.

Below is a second structure, referred to as the flexibly coupled BCFL (FlC-BCFL):
\begin{center}
    \includegraphics[width=0.4\textwidth]{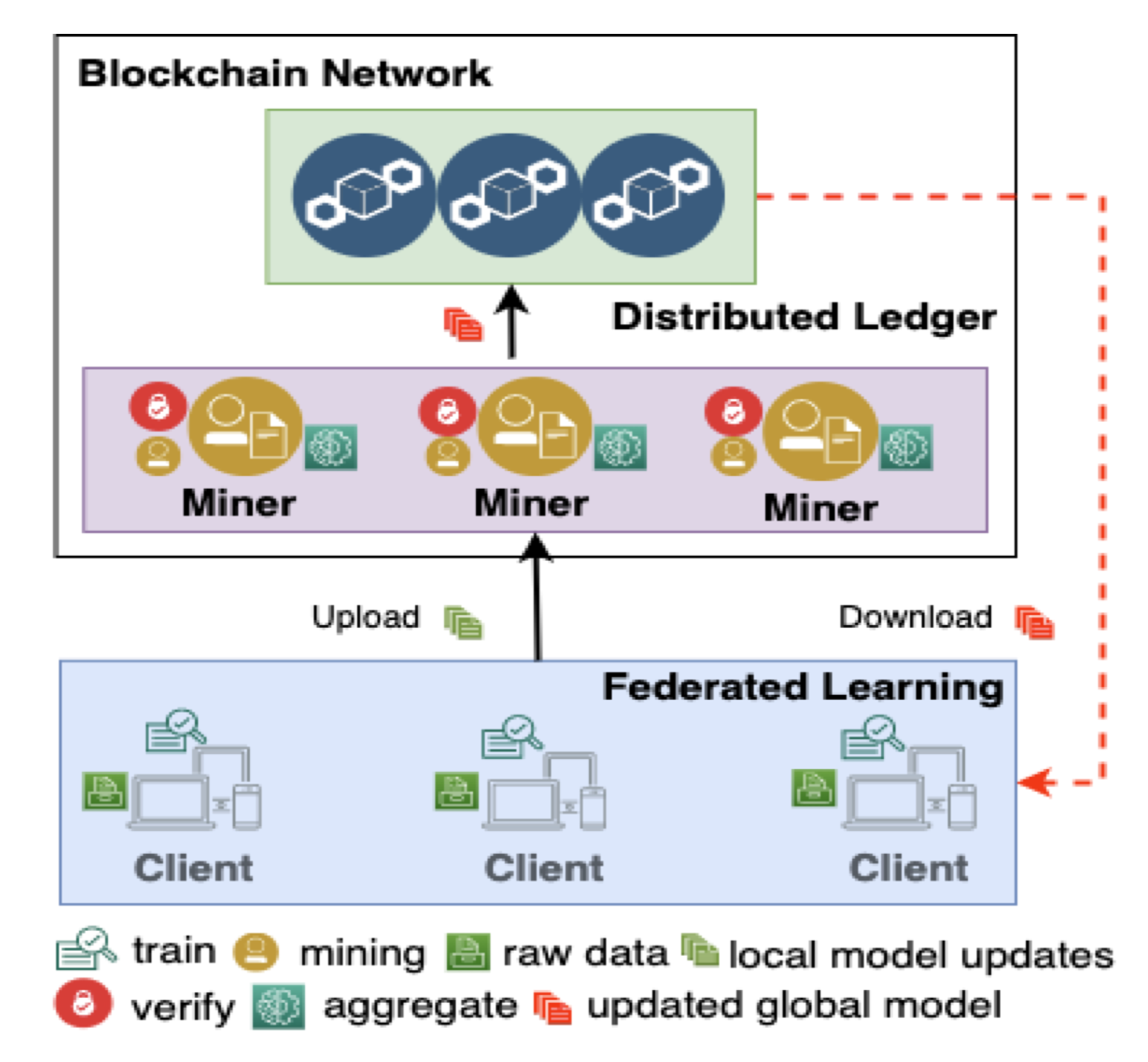}
\end{center}

In this model, the clients and the nodes are separated from each other, so that the weights uploaded by the clients have to be verified by the miners with some verification protocol, and only allowed updates are then pushed to the distributed BC ledger. In a way, we could view the miners together as the aggregator that is fully decentralized, and all other structures are quite similar to the traditional FL.

Finally, it's the loosely coupled BCFL (LoC-BCFL):
\begin{center}
    \includegraphics[width=0.37\textwidth]{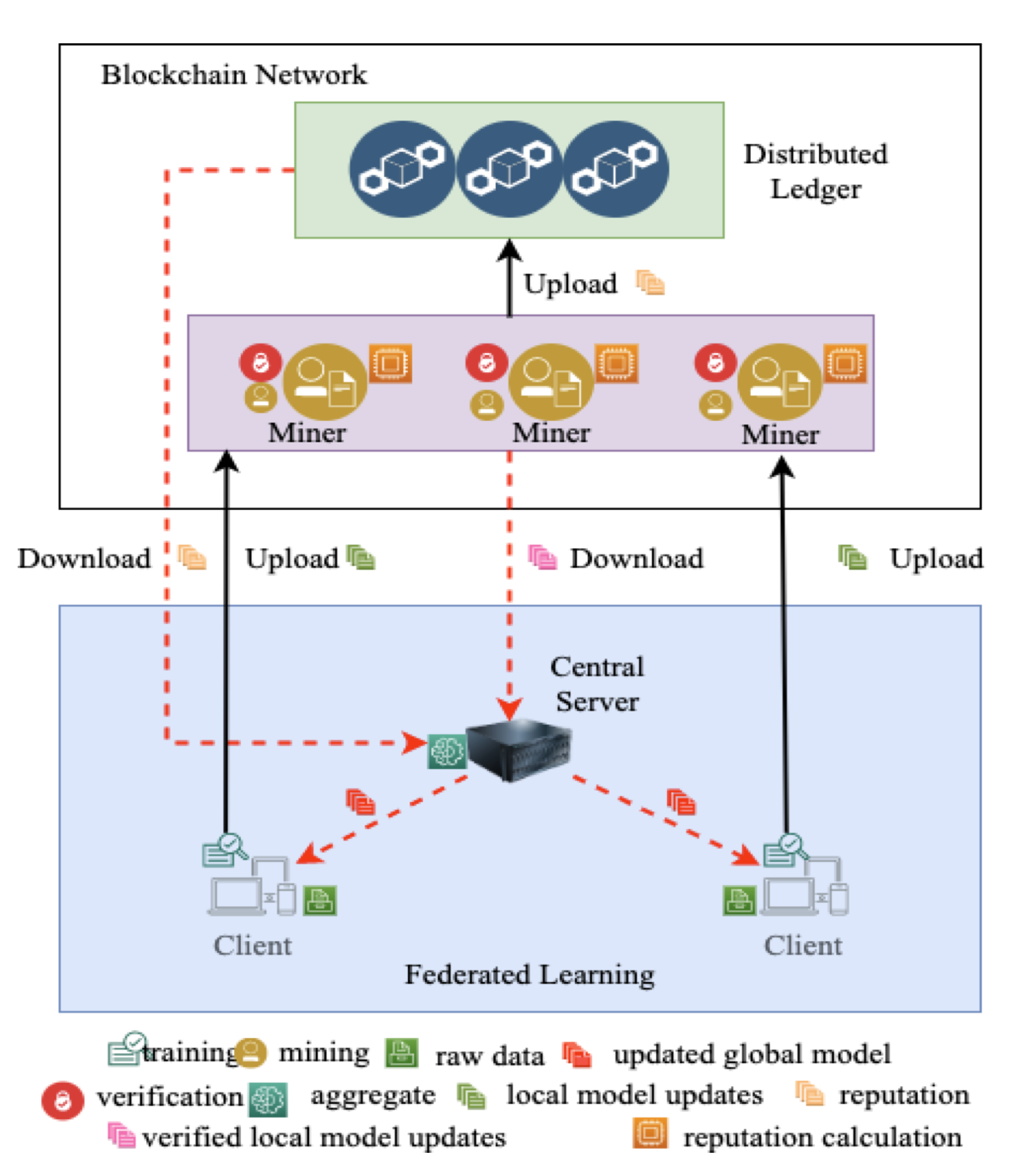}
\end{center}

Here, after clients upload their local model updates, the miners run a reputation algorithm to determine the trustworthy score of the client and the corresponding data. The miner nodes then compete to generate the ledger block. Finally, a central server aggregate the already validated and ranked updates. This is the most loosely coupled structure (thus the name), in that clients and miners do the exact same things as they would in traditional FL and BC respectively. The distributed ledger here only serves the role as a decentralized trustworthy database for all the update of the mutually trained model. With the addition of reputation calculation, this structure is the most robust one against malicious attacks. However, obviously, it also suffers from single point of failure. On the merit side, FL and BC are more disentangled, meaning that clients could completely retain their local data to themselves.

All the above structures are just a qualitative description of the potential categories researcher could approach about this topic of fully decentralized FL, using the existing convenient structure of BC. We now explore a more concrete algorithm that could act as an aggregator, but could be fully decentralized, and therefore be used in nodes in FuC-BCFL and FlC-BCFL.


\subsection{Peer to Peer Federated Learning}

One of the most interesting areas of active research in the field of Federated Learning is in the development of frameworks for Peer to Peer Federated Learning. Lalitha et al. \cite{C} propose a method for Peer to Peer Federated Learning in their 2019 paper. Inspired by social learning theories like De Groot’s model, it proposes each node updating its belief vector $b$ between model training updates by performing a weighted average with its one-step neighbors according to a trust matrix $W$ to get the new belief vector $\rho$ at the $k$th iteration having parameter $\theta$:
\begin{equation*}
    \rho^{(k)}(\theta) = \frac{\exp(\sum_{j=1}^{N} W_{ij}\log b_{j}^{k}(\theta))}{(\sum_{\psi \epsilon \Theta} \exp(\sum_{j=1}^{N} W_{ij}\log b_{j}^{k}(\psi)))}
\end{equation*}

They apply this method to two common machine learning problems: linear regression and a deep neural network. In the case of more classical machine learning algorithms like linear regression, their method is solidly grounded in Bayesian statistics, as they perform updates on the prior and obtain the posterior distribution. However, in the training of neural networks, they acknowledge the prohibitive computational complexity of performing a true local Bayesian update. Hence, they use variational inference to estimate the posterior distribution. 

Lalitha et al. \cite{C} apply their Federated learning algorithm to train a basic fully connected neural network classifier on the MNIST handwritten digit recognition dataset split across two nodes with mixed results. On the one hand, they present interesting methodologies, but on the other, there are obvious problems with their implementation as it exists in the paper. First, the use of only two nodes is in no way similar to the deployment of federated learning in real-world systems, which often consist of thousands of nodes. Second, the weights used for the weighted average performed by each node with its one step neighbors are modeled as hyperparameters, and for $n$ nodes, this would result in $n^{2}$ parameters. Our group aims to develop methods to learn these weights in the trust matrix.

Another development in the context of peer to peer sharing of model training information is explained in the GossipGrad paper by Daily et al. \cite{E} Their paper proposes a method of training models on multiple GPU nodes within the same computing cluster through the propagation of gradients through the network via the gossip protocol. 

\begin{center}
    \includegraphics[width=0.37\textwidth]{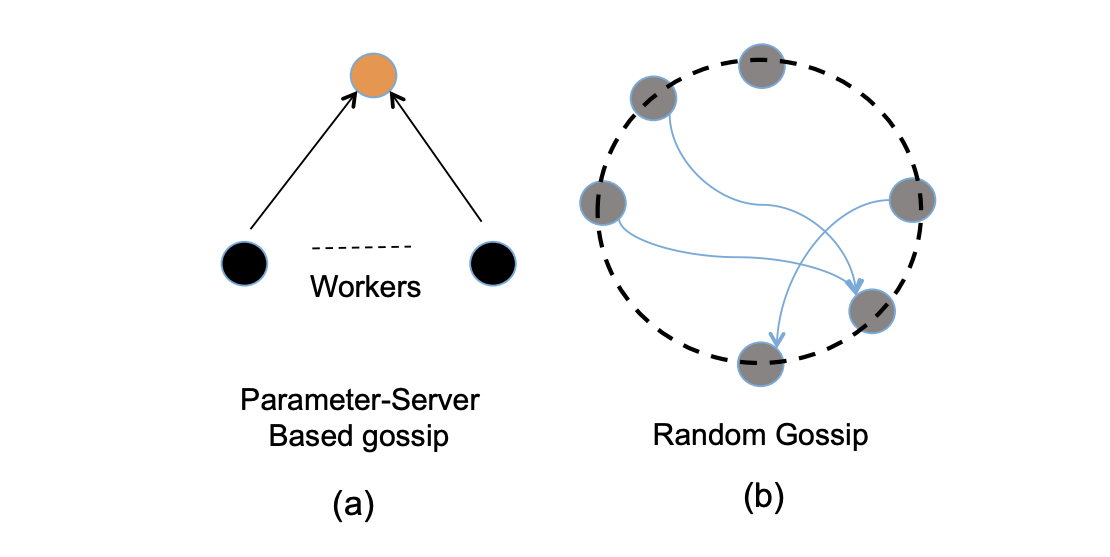}
\end{center}

They achieve commendable results when training larger models than those used in the examples in Lalitha et al. like ResNet and GoogleNet. While this idea is not directly related to federated learning, our group aims to use the ideas of gossip based communication and gradient sharing to improve federated model training.

\section{Project Setup: \href{https://rammkripa.github.io/papaya/}{Papaya}}
\subsection{Peer to peer learning}

To solve the problem of limited privacy and the need for a centralized server in classic federated learning systems, we propose a \href{https://rammkripa.github.io/papaya/}{peer to peer federated learning system}. We developed a system whereby individual nodes on the network store a portion of the data, and each of the nodes trains an instance of a machine learning model (with specific emphasis on deep neural networks) with periodic sharing of parameters between a random subset of nodes according to a learned trust matrix, similar to the classic setting of social learning using DeGroot's model.

\begin{center}
    \includegraphics[width=0.4\textwidth]{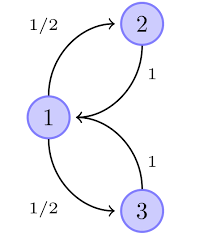}
\end{center}

\subsection{System Architecture}

When building the real-world implementation of the peer to peer learning system, we used an architecture similar to BitTorrent. BitTorrent has two protocols for handling torrents: one that uses a centralized tracker server, and one that uses a distributed hash table without any centralization. We used the hash table based implementation for Papaya. An entity may start a peer to peer learning session by creating a torrent file containing data and model specifications, as well as initializing a bootstrap node if that does not exist, and distributing that file among many clients using a web server or some other method. Clients who take part in the peer to peer learning system would download the torrent file, connect to the distributed hash table network, and begin the training process. \\

Hence each peer obtains a list of all clients taking part in the peer to peer learning process through the hash table. The clients train independently for a few epochs, and then upload their parameters to the distributed hash table. Next, they choose a subset of the available peers from the hash table and train the partner weights. They then perform the weighted average with the chosen partners, and the process repeats itself. \\

\begin{center}
    \includegraphics[width=0.4\textwidth]{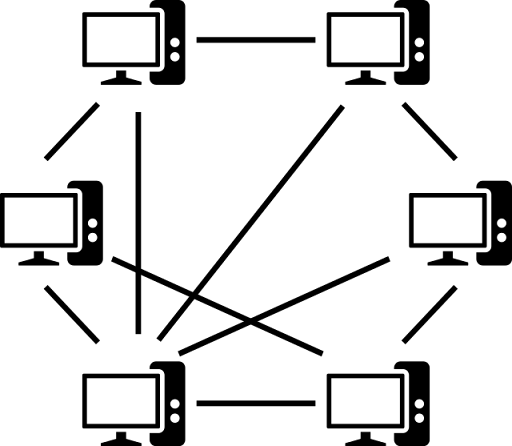}
\end{center}

The peers communicate their model updates to their random partners using the distributed hash table, and then continue the training process. After all of the training epochs are complete, each node has their own version of a fitted model for the data, including the information embedded in other random peers' database as well.  \\

In this way, we anticipate that the bandwidth required by each peer will be minimal, and while a malicious entity could obtain model updates from the distributed hash table, we anticipate that privacy mechanisms will be built in determining access to the distributed hash table. We may also implement differential privacy mechanisms while communicating model updates to reduce the possibility of information leakage. \\

To build this system, we created a framework for clients (nodes in the peer to peer learning process) and a mechanism for the clients to communicate with each other through the distributed hash table. Code for the same can be found here for the \href{https://github.com/rammkripa/papayadhttesting/blob/main/nodestart.py}{client interaction with the hash table} and \href{https://github.com/rammkripa/papayadhttesting/blob/main/papayaclientdistributed.py}{the client itself}. In order to keep consistent with the general distributed hash table implementations, we keep a bootstrap node/server that is always online for the network. This bootstrap server does not act as a computing node; it only serves as a backbone access point for the algorithm to work properly. \\

Some options for the Distributed Hash Table for communication between the clients included BitTorrent distributed hash table (btdht), Hazel, and Kademlia, among others. We settled on a Kademlia-based implementation.



\section{Experiments}

\begin{table}[H]
\vspace{10pt}
\begin{center}
{
\setlength\tabcolsep{10pt}%
\begin{tabular}{|l|l|l|}
\hline
 & Avg Accuracy & std \\\hline
iid balanced & 92.0 & 0.03 \\\hline
non-iid balanced & 91.1 & 0.08   \\\hline
iid non-balanced & 92.1 & 0.04  \\\hline
\end{tabular}}
\vspace{5pt}
\caption{MNIST experiments comparing different setups of data for peer-to-peer distributed learning. Non-iid is when different nodes have different distribution of training labels. Non-balanced is when different nodes have different proportion of training data.}
\label{tab:analysis_results}
\end{center}
\end{table}

\begin{table}[H]
\vspace{10pt}
\begin{center}
{
\setlength\tabcolsep{10pt}%
\begin{tabular}{|l|l|l|}
\hline
 & Avg Accuracy & std \\\hline
iid balanced & 84.2 & 0.10 \\\hline
non-iid balanced & 82.7 & 0.07   \\\hline
iid non-balanced & 83.4 & 0.12  \\\hline
\end{tabular}}
\vspace{5pt}
\caption{Fashion-MNIST experiments with the same setup as above. Using a linear logistic regression model, we showed the degradation due to non-iid or unbalanced data to be minimal.}
\label{tab:analysis_results}
\end{center}
\end{table}

\begin{table}[H]
\vspace{10pt}
\begin{center}
{
\setlength\tabcolsep{10pt}%
\begin{tabular}{|l|l|l|}
\hline
 & Avg Accuracy & std \\\hline
No Sharing & 83.2 & 0.20 \\\hline
Fixed Sharing & 84.1 & 0.10   \\\hline
Learned Sharing & 84.2 & 0.10  \\\hline
\end{tabular}}
\vspace{5pt}
\caption{Fashion-MNIST experiments comparing different setups of parameter sharing scheme for peer-to-peer distributed learning. No sharing is when the nodes are trained separately. Fixed sharing is when the sharing weights are pre-defined and equal. Learned sharing is when the sharing weights are initialized from fixed sharing and learned during training.}
\label{tab:analysis_results}
\end{center}
\end{table}

We followed the same general process to evaluate our results for multiple models and splits of the corresponding dataset. We tested linear models such as logistic regression and more complex models such as one layer neural network and one layer convolution neural network. We chose MNIST and Fashion MNIST as our dataset which was split among 6 virtual training nodes, and evaluated against a held out test set. All models were trained with SGD and a batch size of $500$ for a total of $100$ epochs. Weight averaging took place every $5$ epochs by pairing random partners. To analyze the experiment results, we considered accuracy score and its standard deviation.\\

Regarding the distribution of datasets, we divided the data into two mutually exclusive sets, one for training and one for testing. Within the training dataset, we would further divide the set into 2 subsets of training dataset and validation dataset, having a ratio of approximately 80\%:20\%. In addition to IID balanced split of data between nodes, we will also consider non-balanced IID data and balanced non-IID data splits between nodes, and evaluate performance on these splits. Non-balanced split refers to each node having its own distribution of labels which are deliberately made to be unbalanced. Non-iid split refers to each node having different number of training data points. In both Table 1 and 2, we observe very minimal performance degradation for the harder and more realistic settings (non-balanced and non-iid) which suggests the feasibility of our decentralized parameter sharing approach.\\

\begin{table}[H]
\vspace{10pt}
\begin{center}
{
\setlength\tabcolsep{10pt}%
\begin{tabular}{|l|l|l|}
\hline
 & Avg Accuracy & std \\\hline
iid balanced & 81.0 & 0.13 \\\hline
non-iid balanced & 79.9 & 0.21   \\\hline
iid non-balanced & 80.9 & 0.31  \\\hline
\end{tabular}}
\vspace{5pt}
\caption{Fashion-MNIST experiments with one layer fully connected neural network. We show that with this different setup, we still achieve minimum performance degradation.}
\label{tab:analysis_results}
\end{center}
\end{table}

\begin{table}[H]
\vspace{10pt}
\begin{center}
{
\setlength\tabcolsep{10pt}%
\begin{tabular}{|l|l|l|}
\hline
 & Avg Accuracy & std \\\hline
iid balanced & 82.3 & 0.22 \\\hline
non-iid balanced & 80.7 & 0.13   \\\hline
iid non-balanced & 81.1 & 0.98  \\\hline
\end{tabular}}
\vspace{5pt}
\caption{Fashion-MNIST experiments with one layer convnet. This outperforms the one layer fully connected network. The same trend still holds. But we show that more expressive networks can be used under our framework with increase in computation linear to the model size.}
\label{tab:analysis_results}
\end{center}
\end{table}

We further tested the robustness of our model by applying different neural layer to the last layer of our neural network. In particular, we used traditional linear neural layer and convolution neural layer. Table 3 and Table 4 show the experimental results respectively. If we compare the results to the results in Table 2, we have some interesting observation. First of all, the average accuracy of both using NN layer and CNN layer achieve more than 80\%, which shows the feasibility of our model under different settings. These two settings are more expressive and require more parameters as the additional layer is added. However, we can see that both table 3 and table 4 show a slight drop in average accuracy compared to Table 2 and an increase in standard deviation. There may be many reasons that cause this difference, but we suspect the extra layer may cause our model overfit, which may explain the high variance in the results. In general, the results are aligned with our expectations. We do expect more expressive models are feasible in implementing federated learning. The drop in accuracy does surprise us but many small experiment design choices may also cause such a small drop in accuracy as well. \\

To implement the learned parameter sharing, we allow gradient to flow through the weights and back-propagate all the way during training to update them. Table 5 shows that performance gets better when there’s parameter sharing between nodes which allows for additional data to be shared across the entire network which is embedded in the model parameters. The results also suggest that a learned approach for the trust matrix could be more desirable than equal sharing between all nodes. \\

\section{Discussion}

For the peer to peer learning system, we anticipate that privacy, scalability, and bandwidth requirements will need to be tested. As of now, this peer-to-peer learning system is still at its early stage of rudimentary implementation using the Kademlia distributed hash table. While translating this framework into a real world deployable system, we must ensure that we can offer similar privacy guarantees and scalability as federated learning models. We anticipate that our solution will require less bandwidth at each peer than at the centralized server in a federated learning framework because model updates are not being aggregated at every node. Bandwidth required at each peer may prove to be a complication, in which case compression and protocols limiting the number of parameters to exchange can be used to alleviate the issue. \\

\section{Conclusion}

Traditional Federated Learning requires a centralized server. This exposes the system to a single point of failure, raising privacy and security concerns. To tackle this, we implemented fully decentralized Federated Learning where model updates only happen within a node or through communication between random node pairings. With image classification datasets, we showed empirically that this setup could eliminate the single point of failure while maintaining model performance.

We leave improving our existing system to future work. The communication channel in our infrastructure can be made more scalable and secure by incorporating stronger verification schemes. In terms of the learning algorithm, instead of doing simple arithmetic averaging between nodes, more information like demographic similarity could be used to customize the weight sharing procedure.

\newpage
\raggedright

\bibliographystyle{plain}
\bibliography{\jobname}
\end{document}